\documentclass[conference]{IEEEtran}
\IEEEoverridecommandlockouts
\usepackage{algorithmic}
\usepackage{graphicx}
\usepackage{textcomp}
\usepackage{cite}
\usepackage{booktabs}
\usepackage[dvipsnames]{xcolor}
\usepackage{listings}
\usepackage{todonotes}
\usepackage{hyperref}

\newcommand\YAMLcolonstyle{\color{red}\mdseries}
\newcommand\YAMLkeystyle{\color{black}\bfseries}
\newcommand\YAMLvaluestyle{\color{blue}\mdseries}
\makeatletter
\newcommand\language@yaml{yaml}
\expandafter\expandafter\expandafter\lstdefinelanguage
\expandafter{\language@yaml}
{
  keywords={true,false,null,y,n},
  keywordstyle=\color{darkgray}\bfseries,
  basicstyle=\YAMLkeystyle,
  sensitive=false,
  comment=[l]{\#},
  morecomment=[s]{/*}{*/},
  commentstyle=\color{purple}\ttfamily,
  stringstyle=\YAMLvaluestyle\ttfamily,
  moredelim=[l][\color{orange}]{\&},
  moredelim=[l][\color{magenta}]{*},
  moredelim=**[il][\YAMLcolonstyle{:}\YAMLvaluestyle]{:},  
  morestring=[b]',
  morestring=[b]",
  literate =    {---}{{\ProcessThreeDashes}}3
                {>}{{\textcolor{red}\textgreater}}1     
                {|}{{\textcolor{red}\textbar}}1 
                {\ -\ }{{\mdseries\ -\ }}3,
}
\begin{document}

\title{Evaluation of Sampling-Based Optimizing Planners for Outdoor Robot Navigation 
}

\author{\IEEEauthorblockN{1\textsuperscript{st} Fetullah Atas}
\IEEEauthorblockA{\textit{Faculty of Science and Technology } \\
\textit
{
Norwegian University of Life Sciences}\\
Aas, Norway \\
fetullah.atas@nmbu.no
}
\and
\IEEEauthorblockN{2\textsuperscript{nd} Lars Grimstad}
\IEEEauthorblockA{\textit{Faculty of Science and Technology } \\
\textit{
Norwegian University of Life Sciences} \\
Aas, Norway \\
lars.grimstad@nmbu.no
}
\and
\IEEEauthorblockN{3\textsuperscript{rd} Grzegorz Cielniak}
\IEEEauthorblockA{\textit{Lincoln Centre for Autonomous Systems} \\
\textit
{
University of Lincoln}\\
Lincoln, United Kingdom \\
gcielniak@lincoln.ac.uk}
}

\maketitle

\begin{abstract}
Sampling-Based Optimal(SBO) path planning has been mainly used for robotic arm manipulation tasks. Several research works have been carried out in order to evaluate performances of various SBO planners for arm manipulation. However, not much of work is available that highlights performances of SBO planners in context of mobile robot navigation in outdoor 3D environments. This paper evaluates performances of major SBO planners in Open Motion Planning Library(OMPL) for that purpose. 
Due to large number of existing SBO planners, experimenting and
selecting a proper planner for a planning problem can be burdensome and ambiguous. SBO planner's probabilistic nature can also add a bias to this procedure. 
To address this, we evaluate performances of all available SBO planners in OMPL with a randomized planning problem generation method iteratively. Evaluations are done in various state spaces suiting for different differential constraints of mobile robots. The planning setups are focused for navigation of mobile robots in outdoor environments. The outdoor environment representation is done with prebuilt OctoMaps, collision checks are performed between a 3D box representing robot body and OctoMap for validation of sampled states.
Several evaluation metrics such as resulting path's length, smoothness and status of acquired final solutions are selected. According to selected metrics, performances from different SBO planners are presented comparatively.
Experimental results shows the significance of parallel computing towards quicker convergence rates for optimal solutions. Several SBO methods that takes advantage of parallel computing produced better results consistently in all state spaces for different planning inquiries.
\end{abstract}


\section{Introduction} \label{introduction}

Path planning is one of the essential steps for navigating a robot from its current pose to a
desired pose in space while avoiding obstacles. 
Various approaches to solve planning problem has been purposed over the years. 
Sampling-Based Optimal(SBO) planning is one of these 
approaches that got attention of researchers in recent years due to its effectiveness while 
planning in complex, high dimensional state spaces. 
SBO planners aims to minimize or maximize an objective function, 
the ultimate goal is to get a valid plan with the lowest cost or highest award depending on the selected objective function.
In SBO planning, the objective function is mostly a cost function, where the amount of cost can be
dependent on different properties of the path such as shortness, smoothness or estimated energy consumption along path.

\begin{figure}
  \includegraphics[width=\linewidth]{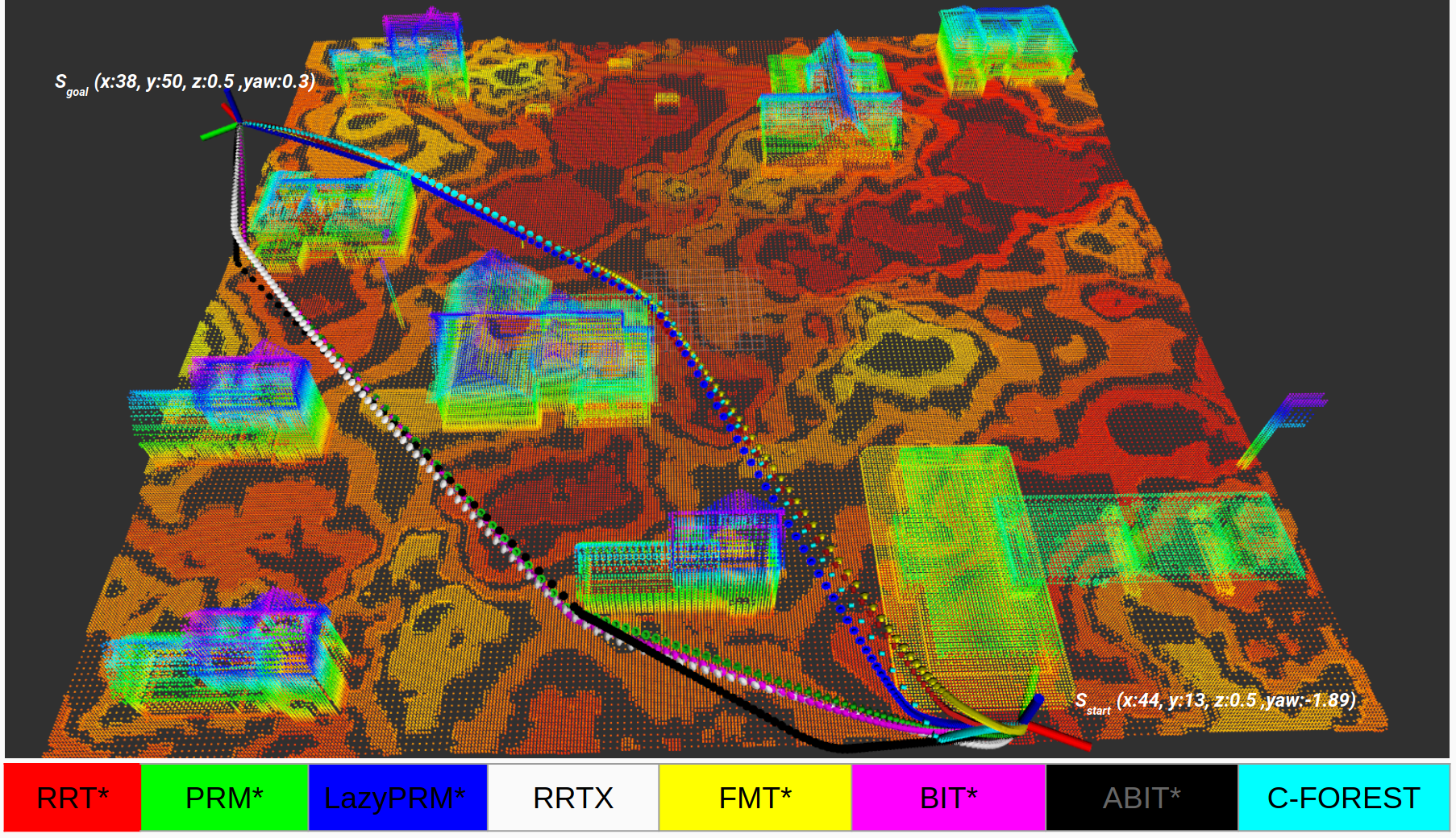}
  \caption{Sample planning results from evaluated SBO planners in SE3 state space, the ground is uneven hence the contour like shapes}
  \label{fig:se3_result}
\end{figure}

The path planning problem is well studied and established for the most part. However,
due to different topological structures and scales of environments, 
increased dimensions of states for different types of robots, 
path planning is still a relevant research area particularly in robotics.
There isn't one general method or framework that can cope with all different planning problems,
different groups of planners are better suited to different situations. Therefore it is important to systematically evaluate planners for mobile robots rigorously. Lack of work in literature that focuses SBO path planning performance analysis for outdoor robot path planning has been main motivation behind this study. Some of the main contributions of this work are listed as follows.
\begin{itemize}
    \item To the best of our knowledge, we are first to present a detailed evaluation of SBO planners that particularly focuses on outdoor robot navigation, with realistic planning times towards actual application. Evaluations are carried out for robots with different kinematic constrains. 
    \item We present comparative results from all major SBO planners with randomized iterative approach that can better capture behaviour of probabilistic SBO planners.
   \item The environment representation enables to analyze response of SBO planners for navigation in 3D, plans can go through light ramps, slopes and elevations within defined boundaries.
\end{itemize}

This paper is a result of an empirical study of state-of-the-art SBO planners and their behaviours under different state spaces. 
The evaluations are performed in different state-spaces suiting for all well known robot kinematics. In section a literature summary of path planning is provided.
The section \ref{sbo_overview} gives a brief overview of SBO planners subject to evaluation, in section \ref{experiments} 
the metrics and setup for evaluation are described.
Finally, in section \ref{discussion_results} a discussion related to acquired results and in section
\ref{conclusion} conclusions drawn from this study are provided.  

\section{Related Work} \label{related_work}

SBO planning methods 
are extended on top of basic path planning algorithms into more challenging scenarios and diverse
applications \cite{6722915}. 
Complete path planning algorithms existed as early as 1979 \cite{LIS3}, but the complexity
of the these approaches made it hard for practical use. With arrival of cell 
decomposition methods such as Brooks and Lozano-Perez \cite{article} and Potential Fields methods 
\cite{doi:10.1177/027836498600500106}
the requirement of completeness was relaxed to \textit{resolution completeness.}, that is
the algorithm will return a valid path if one exists with a finely set resolution for the 
decomposed cells representing environment.
These algorithms proved to work well in practical use cases \cite{1020564024509}. However, for the
problems with higher number of dimensions and scales, cell decomposition based methods suffered 
due to increasing number of cells which caused impractical computation times. 
Cell decomposition methods used a complete representation of the 
environment it operated in. A predefined complete representation of environment resulted in extensive computation time,
making these methods unfeasible for high dimensional state spaces.

This led to development of Sampling-Based(SB) planners such as Kavraki and Latombe \cite{350966}, Kavraki et al. \cite{508439}
and LaValle \cite{Lavalle98rapidly-exploringrandom}. 
Instead of relying on full representation of the environment, SB methods samples a series of states while performing a collision check for each of this state to ensure validity of generated states. The way each of
SB planner samples a state differs. The probabilistic characteristic of these methods contradicts with
completeness of these methods. However, SB planners are \textit{probabilistically complete} \cite{508439} meaning that,
if a valid solution exists the planner will return one 
as the number of sampled states approaches to infinity \cite{doi:10.1177/027836499701600604}. Initial SB planners were not cost-aware, however with introduction of novel methods by Karaman and Frazzoli 2011 \cite{karaman_frazolli}, a new milestone that led to numerous research efforts on SBO planners was reached. In next section we enhance our discussion on SBO planning in next section, each SBO planner subject to our evaluation is highlighted with its unique property.

\section{Sampling-Based Optimal Planners} \label{sbo_overview}

Planning is a core component of autonomous robot's software stack alongside Perception and Control. It is a process of estimating series of consecutive valid, collision-free states(or poses) that will drive the robot to a desired pose safely. Here, SBO approach might be particularly useful while planning in mid-scale outdoor environments(e.g. 100 x 100 meters) with 3D structures such as ramps, elevations that cant be represented in 2D, hence inconvenient to use 2D grid cell based methods for such environments. In following we take a close look at existing SBO Methods and their characteristics.

Two of well established Sampling-Based(SB) planners are Rapidly-Exploring Random Trees(RRTs) \cite{Lavalle98rapidly-exploringrandom},
and Probabilistic RoadMaps(PRMs) \cite{508439}. Although employing some common structures RRTs and PRMs
differs from each other in the way they construct a graph(or tree for RRT) for connecting start \(S_{start}\) to goal \(S_{goal}\) pose.
A major difference of PRM over RRT is that it can be used as multiple-query planner, while RRT is suited for single-query online planning. 
Depending on characteristic of the environment this can be useful or vice-versa. 
A robot that operates in a highly dynamic environment, may not benefit much from the multiple-query feature of PRMs, 
whereas an online single-query RRT could be practical. 

Generating a valid plan with an SB planner is possible(if one exists) most of the time.
Several variants of RRTs has been deployed to robotic platforms \cite{Branicky}, \cite{doi:10.1177/0278364910388315},
and it has been proven useful for practical use cases. In many cases though, the 
quality of this valid paths are crucial to the overall behaviour of a system. 
Therefore, it is important to systematically define and evaluate quality of a generated path. SBO path planning tries to address this by using various optimization techniques. 

On the high level, SBO planners operates similarly, that is by randomly sampling a
series of states in the state space, making sure that generated states are valid(collision-free). 
A path is then constructed by connecting start state \(S_{start}\) to a goal state \(S_{goal}\) through 
sampled valid states with the minimal cost. Below we list some common procedures that a typical SBO planner employs.

\begin{itemize}
  \item \textit{Sampling:} It is a process of generating sample states in the state space, the randomized state generation has proved to
                          greatly simplify planning in high dimensional state spaces.
  \item \textit{Objective cost calculation:} At this step, the aim is to find a numerical cost of taking robot from a state \(S_a\) to 
                  \(S_b\). The calculation of this cost will depend on the defined objective, e.g. for path length, this cost can be Euclidean distance between
                  \(S_a\) and \(S_b\).
  \item \textit{Collision checking:} It is mostly a Boolean function, which checks whether a defined robot body volume is in collision with its environment at a sampled state. This step takes most part of execution time \cite{1438016}.  
  \item \textit{Nearest neighbor search:} Based on selected objective cost, this procedure calculates nearest neighbor or minimal cost to a newly generated
                                          sample.                                                                                       
\end{itemize}

Some initial works towards SBO planning relied on different heuristics, e.g. by using a biased model to grow RRTs into low-cost regions \cite{1248805}.
Berenson et al. introduced Transition-RRT(T-RRT), which was designed to steer the exploration of RRT by considering stochastic global optimization method.

Karaman and Frazzoli \cite{karaman_frazolli} 
extensively investigated optimality of popular SB methods, RRTs and PRMS.
The analysis concludes that the resulting solutions by these planners
under normal circumstances converges to a non-optimal value almost at all times. Despite this fact,
the authors purposed two new milestone approaches known as RRT* and PRM*.
These two planners are provably asymptotically optimal, meaning that the cost returned by these 
planners converges to optimum value almost at all times. The proof of asymptotic optimality
led to whole another effort in order to improve the quality guarantees of SBO
planners. Recent works focused on improving the convergence rate to optimality and better initial 
solutions. Some essential properties of RRT* and PRM* can be summarized as follows.

\subsection*{RRT*:}
RRT* differs from original implementation in a sense it restructures the tree 
by considering the total cost of that particular sample to the start state, whereas in RRT 
a valid newly sampled state is just added to the nearest node in the tree. This property makes RRT* 
path cost-aware and leads much shorter and smoother paths with a cost of increased computation.
\subsection*{PRM*:}
PRM* does not sample states with a fixed radius as in PRM, instead the radius is set to be a function of state space dimensions and
characteristic of state space. This leads to sparse and quick exploration of large 
collision free areas, as well dense exploration in narrow nooks finally leading increased exploration capabilities.

Other SBO planners subject to evaluation in our work bases on above two with slight 
modifications, we will lightly mention their general characteristics. 
LazyPRM* is a combination of \cite{karaman_frazolli} and \cite{lazyprm}.
Number of nodes to connect are not fixed but is determined automatically 
based on coverage of the state space \cite{6377468}. SPARS  \cite{SPARS} and SPARStwo \cite{SPARStwo} are also built on top of 
PRM*, these methods suggests finite-sized nodes within roadmap which might provide near-optimality properties. If the optimality is relaxed to near-optimality, sparser graphs can provide solutions meeting near-optimality requirement.
This lead to several advantages such as reduced resources used to solve the planning problem and decreased rate of node addition into roadmap.

RRTX \cite{Otte2014RRTXRM} is a
variant of RRT*, the nodes in the tree are rewired only if the cost is being reduced 
by a specified margin(known as \textit{epsilon}), this improves the convergence of solution. 
RRTsharp \cite{RRTsharp}, bases on Rapidly-exploring Random Graphs(RRG), it improves the convergence rate by keeping stationary information of
sampled vertices, by doing this RRTsharp can achieve better estimations of potential optimal paths. 
InfromedRRT* \cite{infromedRRTstar} narrows down tree exploration space to boundaries of a prolate hyperspheroid, a prolate hyperspheroid is n-dimensional symmetric ellipse that is constructed around current optimal plan.
As the search space is reduced to region around optimal plan, the converge rate improves significantly.  
SST \cite{sst}, an enhancement of RRT* tries to address optimal kinodynamic planning when there is no access to two-point boundary value problem(BVP) solver, BVP is required to get desirable property for systems with dynamics. 

A more recent SBO approach, Batch Informed Trees (BIT*) \cite{BIT*} uses a unified structure that is composed of both 
graph and trees, this leads to a few advantages such as being able to use heuristics like one in A* \cite{Hart1968}
to prioritize tree expansions towards the goal with high quality paths. An extension of BIT*, 
Advanced BIT*(ABIT*) \cite{Strub_2020}, combines an advanced graph-based search with an RRT* like planner for faster 
convergence rate. Another variant of BIT*, AIT* \cite{AIT*}, uses an asymmetric bidirectional search where both of the searches continuously inform each other, this leads to quicker initial solutions. 

According to authors \cite{fmt} Fast Marching Tree (FMT*) is an SBO planner that converges to optimal solutions faster than PRM* and RRT*. The approach
utilizes recursive dynamic programming to form a tree from probabilistically sampled states.

Anytime solution optimization by Luna et.al. \cite{aps} is a generic method that wraps around one or more SBO planner. The method applies 
an iterative path simplification, shortening process. In case of multiple SBO planner, it is capable of generating \textit{hybrid} paths if they are to  yield shorter solutions. 
Another approach CFOREST \cite{cforest}, wraps a parallelized framework around single-query SBO planners e.g. RRT*.
Separate trees are grown from different threads while these threads communicate with each other, every time
a better path is found by a thread, other threads are also informed. This leads a significant speedup of convergence towards optimal plans.
The end result paths are shorter and smoother.

\section{Experiments} \label{experiments}

\subsection{Evaluated Planners}
OMPL is an open source C++ library that contains many of state-of-the-art SBO planning algorithms.
Its abstract interface allows for quick deployment of many SB and SBO planners with little effort in terms of lines of code. 
At the time of writing, all available SBO planners in OMPL(see Table \ref{tab:planner_list}) \cite{6377468} were
selected for the evaluation. These includes very recent SBO planners such as AIT* \cite{AIT*} ABIT* \cite{Strub_2020}. All SBO
planners were evaluated with their default settings.
Planners with parallelization capabilities such as 
CFOREST and AnytimePartShortening(APS) were used with 8 instances of RRT* in the background.

\begin{table}[ht]
  \centering
  \tiny
  \begin{tabular}{|c|} \hline
  PRM*             \\ \hline
  LazyPRM*         \\ \hline
  RRT*             \\ \hline
  RRTsharp         \\ \hline
  RRTX             \\ \hline
  InformedRRT*     \\ \hline
  BIT*             \\ \hline
  ABIT*            \\ \hline
  AIT*             \\ \hline
  LBTRRT           \\ \hline
  SST              \\ \hline
  SPARS            \\ \hline
  SPARStwo         \\ \hline
  FMT*              \\ \hline
  CFOREST          \\ \hline
  AnytimePartShortening(APS)  \\ \hline
 \end{tabular}
 \space
 \caption{List of Optimizing planners used for evaluation.}
 \label{tab:planner_list}
\end{table}

\subsection{Scenarios and Evaluation Setup}
For the collision checking, we use Flexible Collision Library(FCL) 
\cite{fcl}.
The environment is represented with an OctoMap \cite{Hornung12octomap}, a memory efficient tree data structure that holds 3D
occupancy information of environment in a probabilistic manner. 
Collision checking is performed between a 3D bounding box that
represents robot body and a prebuilt OctoMap of the environment. 
We evaluate planners in OctoMaps which were generated from Gazebo \cite{gazebo} simulator. 

The simulated environment may consists of an even/uneven ground plane with various objects such as houses, lamb posts and plants in. Figure \ref{fig:se3_result} shows an corresponding OctoMap of such environment with sample results from several SBO planners.

During the experiments, we use a randomized goal and start pose generation within defined state space boundaries. The validity of these start and goal poses were acquired by collision check status
between a 3D box representing robot's body and OctoMap at start and goal poses. In order to make sure that an actual valid plan exists between randomly generated start and goal poses, a planning request was made from start to goal pose with a sufficient amount of timeout(\textbf{10 seconds}). 
The actual benchmarking was initialized 
only if there existed a valid plan from random start to random goal pose.
In Final phase, RRT* planner with a very large amount of timeout(500 seconds) was used to create a ground truth plan, this ground truth plan was used as base for comparison of results from all SBO planners. 

\begin{figure}
  \includegraphics[width=\linewidth]{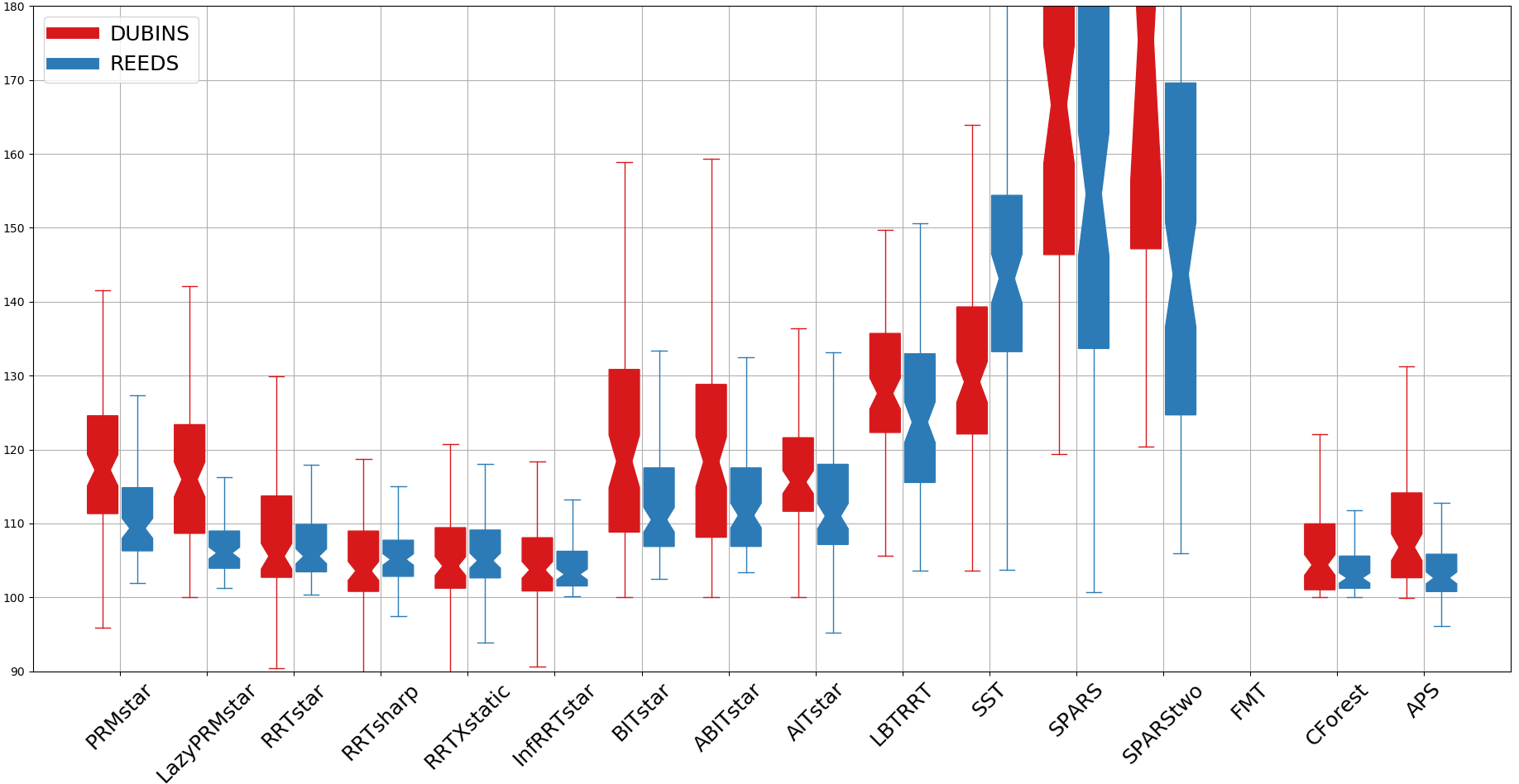}
  \caption{Resulting path lengths in  REEDS and DUBINS spaces, ground truth path lengths are normalized to 100 meters, all results are normalized accordingly}
  \label{fig:reeds_dubins_lengths}
\end{figure}

The length of ground truth plans and resulting plans from SBO planners were normalized to 100 meters as seen in Figures \ref{fig:reeds_dubins_lengths}, \ref{fig:se2_se3_lengths}.
This allows to simpler interpretation of SBO planner's performance for length objective. 

The planners are 
evaluated in various state spaces; SE3, SE2, REEDS-SHEEP \cite{reeds},
DUBINS \cite{dubins}. 
The variety of these state spaces are 
intentionally selected to meet existing kinematic structures of grounds robots, e.g. Differential-Drive, Omni-Drive, Ackermann etc. as well as to confirm consistency of planners under different constrains.
Paths in REEDS-SHEEP \cite{reeds} and DUBINS state spaces particularly suits robots with Ackermann kinematics. We also perform evaluations in SE3 for 
uneven terrains such as one depicted in Figure \ref{fig:se3_result}. 
The benchmark scripts are written in C++ and are interfaced with OMPL.
A YAML file format is used to specify and configure parameters for different state spaces and other parameters benchmarks requires. This allows efficient testing of different planners in various state spaces.
All configuration parameters are highlighted below, description of each parameter is given in red color.

{\tiny
\begin{lstlisting}[language=yaml]
    planner_timeout: 1.0 # amount of time allowed for planner to get a solution 
    interpolation_parameter: 120 # uniformly interpolate the final path
    octomap_topic: "octomap" #  ROS topic to subscribe for an OctoMap
    octomap_voxel_size: 0.2 # cell size for subscribed OctoMap
    selected_state_space: "SE2" # other options are; "DUBINS","REEDS","SE3" 
    selected_planners: ["RRT*","PRMstar", ...] # includes all planners given in Table I.
    min_turning_radius: 1.5 #For DUBINS AND REEDS
    state_space_boundries: # sampling is always within these boundaries
      minx: -100.0
      maxx: 100.0
      miny: -50.0
      maxy: 50.0
      minz: 0.0
      maxz: 2.0
      minyaw: -3.14
      maxyaw: 3.14
    robot_body_dimens: #3D box surrounding robot
      x: 1.5
      y: 1.5
      z: 0.5
    goal_tolerance: 0.2  # in meters
    min_euclidean_dist_start_to_goal: 65.0 # in meters
    epochs: 20 # desired number of valid pairs of start/goal poses
    batch_size: 5 # run each planner this times for a valid pair start/goal pose
    max_memory: 4096 # MB
    results_output_file: "/path_to/result.log"
    visualize_a_sample_benchmark: true 
    sample_benchmark_plans_topic: "benchmark_plan"
\end{lstlisting}
}

\subsection{Evaluation metrics}

There are few important metrics to assess quality of resulting plans. Most essential ones being the length, execution time, availability of
exact solution etc. SBO planner's execution time is quite ambiguous and needs an explanation here, SBO planers uses a timeout parameter, initially
some amount of this time is used to construct a valid plan(if one exists), after a valid plan is found the remaining time is utilized for getting lower cost paths, at the end all time is utilized. Despite being crucial indicator of a planner's performance, we cannot really set execution time as a metric since SBO planners utilize all given time to construct most optimal plan even after a valid plan is found. All planners timeouts were set to \textbf{1 second} for all state spaces. 
Metrics of evaluation and their short descriptions are given as follows.

\begin{itemize}
    \item \textit{status}:  whether a solution(exact or approximate) was found
    \item \textit{solution length}: real length of the resulting solution path
    \item \textit{solution smoothness}: the smoothness of path is calculated by sum of angles between consecutive path segments, the lower this sum the smoother path is. For instance a perfectly straight line will result in smoothness value of 0.0
\end{itemize}

\section{Discussion and Results} \label{discussion_results}
It must be noted that given enough amount of timeouts, each planner's performance will be reasonable and similar. Therefore we used a rather strict timeout amount of \textbf{1 second} to challenge and reveal performance differences among SBO planners. We generated 25 random planning problems with minimum euclidean distance between start and goal pose being \textbf{85 meters}. To reduce bias related to probabilistic nature of planners each planner was inquired to make a plan for each problem \textbf{4 times}, resulting total number 100 runs. The resulting figures are presenting results from 100 runs.

\begin{figure}
  \includegraphics[width=\linewidth]{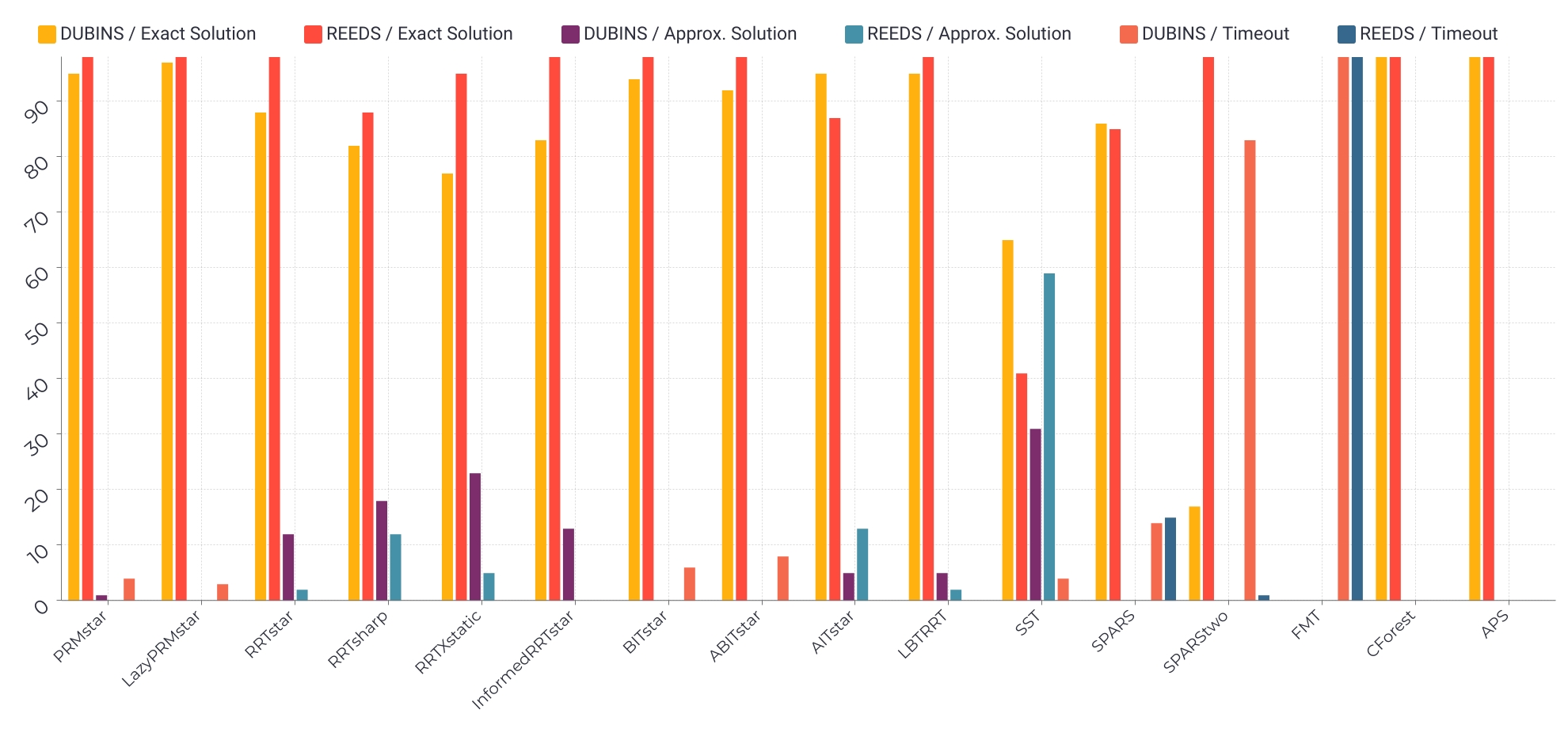}
  \caption{Status of acquired results in REEDS and DUBINS spaces}
  \label{fig:reeds_dubins_status}
\end{figure}

\begin{figure}
  \includegraphics[width=\linewidth]{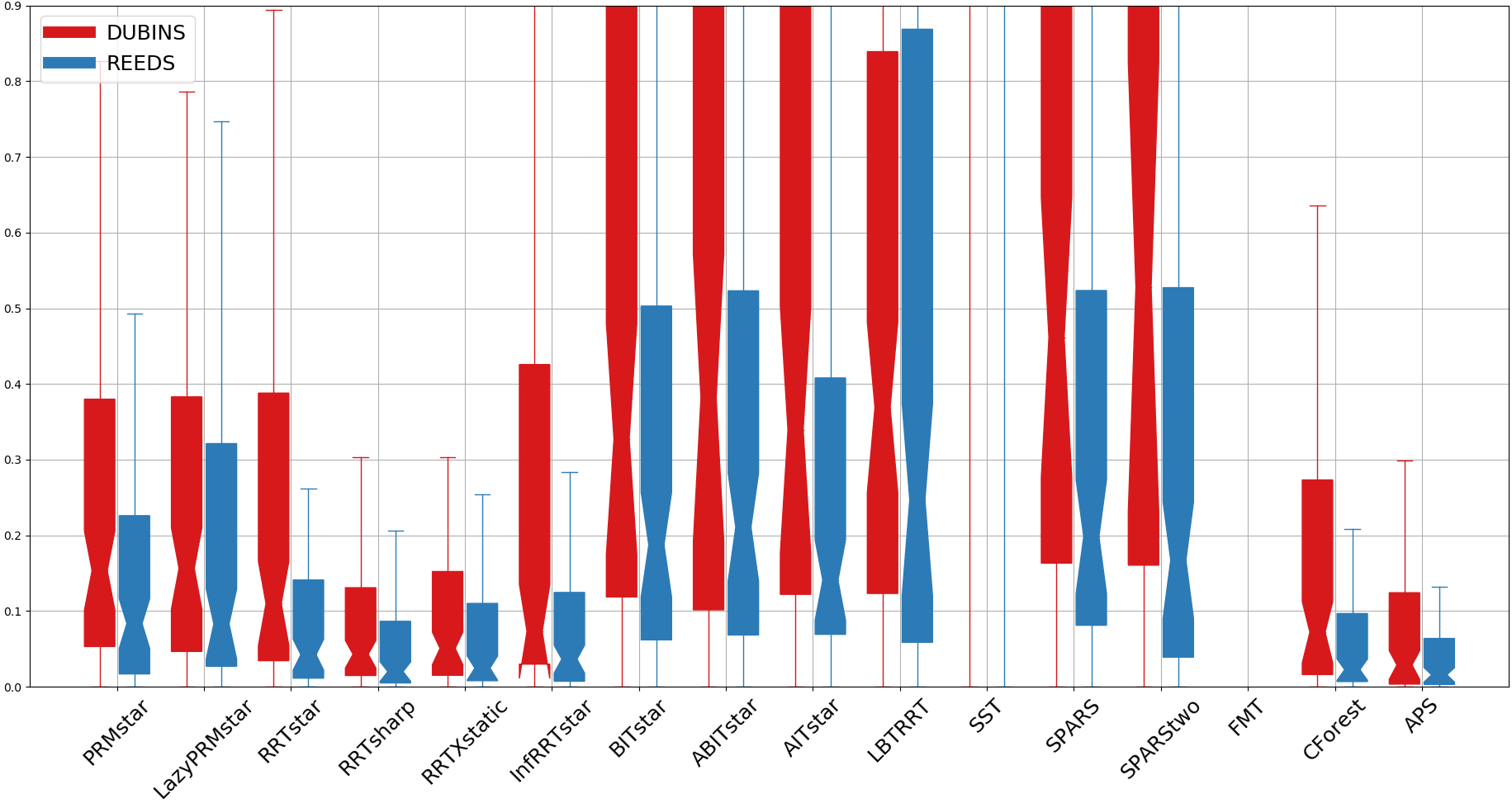}
  \caption{Box plot results of path smoothness in REEDS and DUBINS spaces space}
  \label{fig:reeds_dubins_smoothness}
\end{figure}

\begin{figure}
  \includegraphics[width=\linewidth]{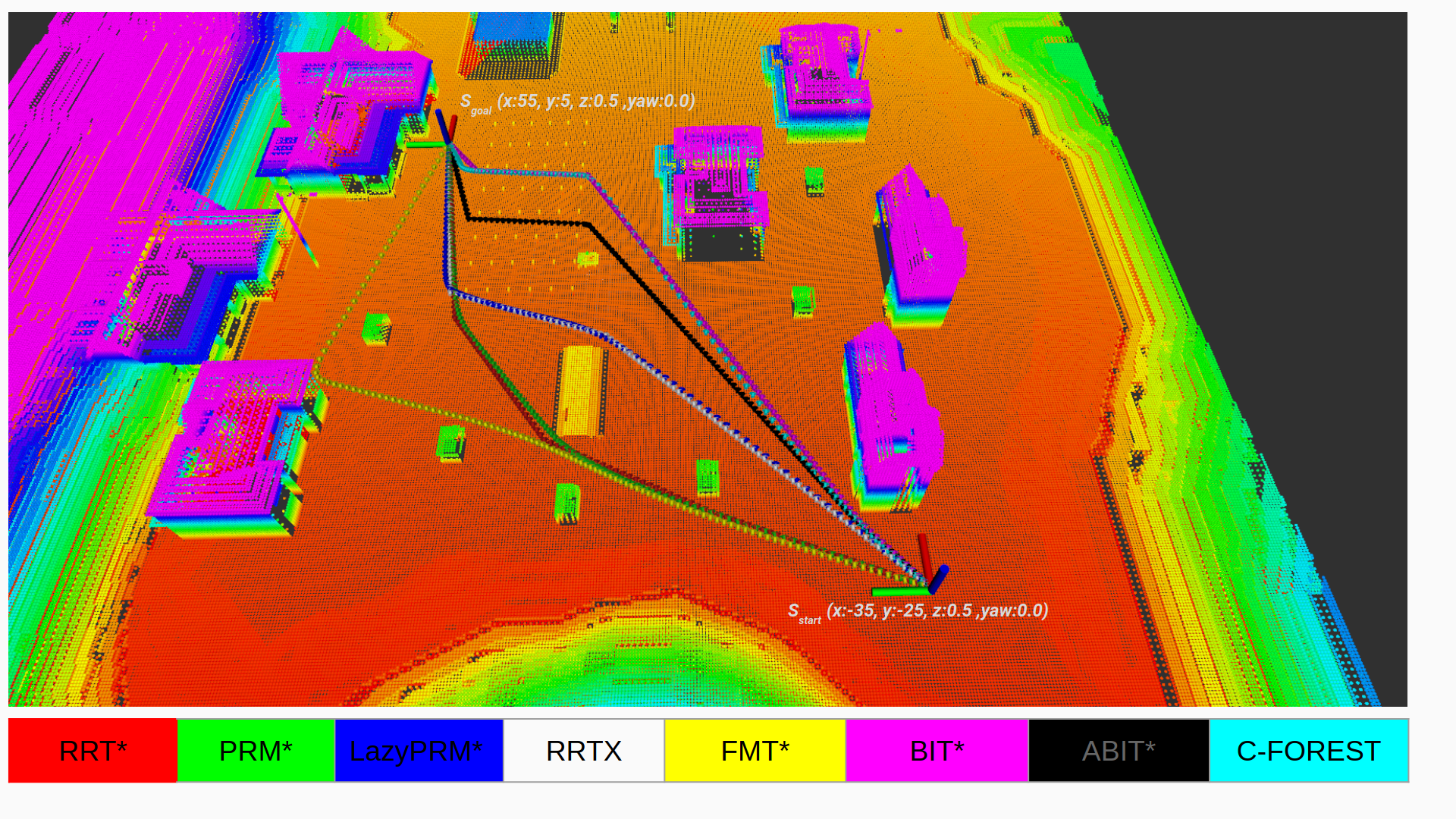}
  \caption{A sample result from SBO planners in SE2 state space}
  \label{fig:sample_result_se2}
\end{figure}

\begin{figure}
  \includegraphics[width=\linewidth]{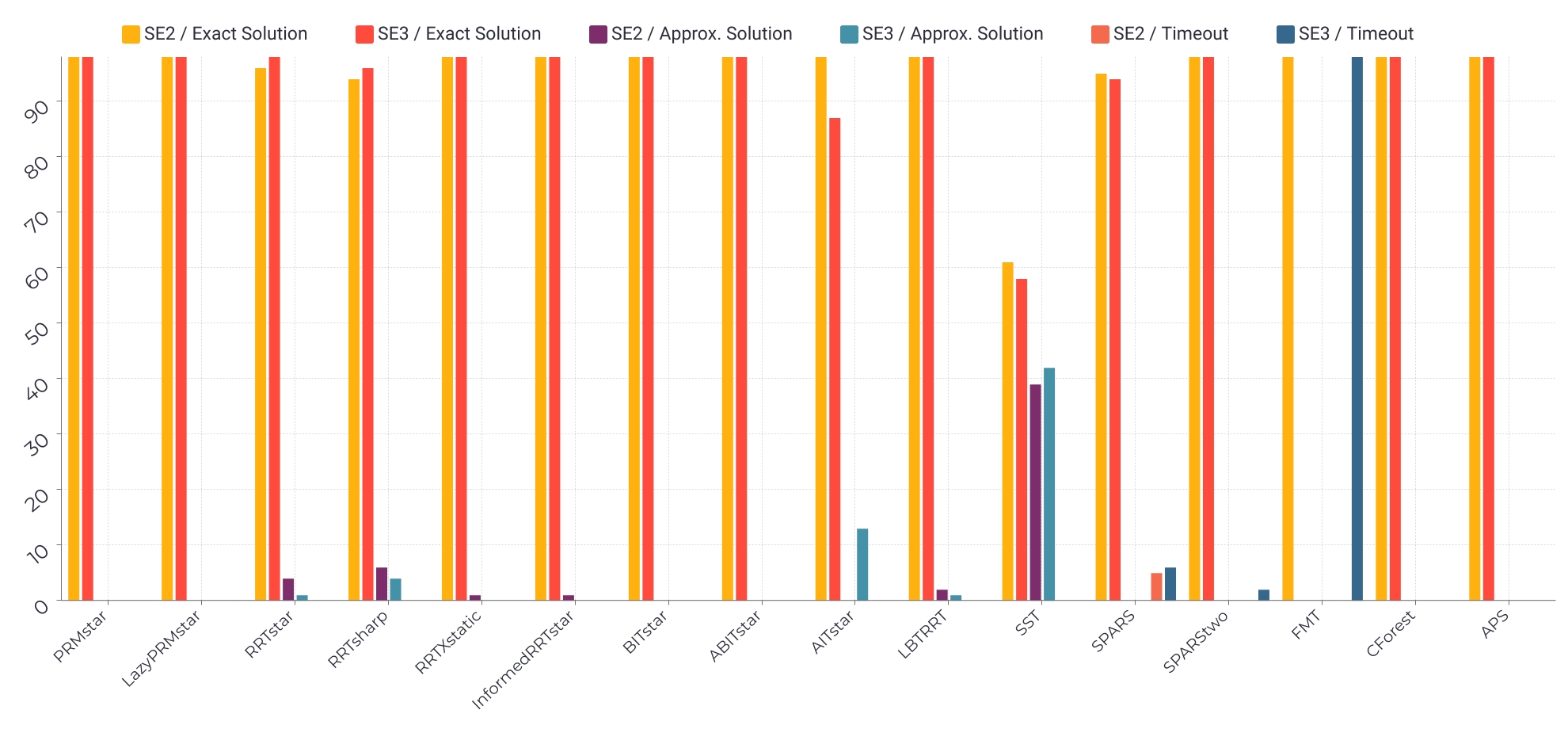}
  \caption{Status of results in SE2 and SE3 spaces}
  \label{fig:se2_se3_status}
\end{figure}

\begin{figure}
  \includegraphics[width=\linewidth]{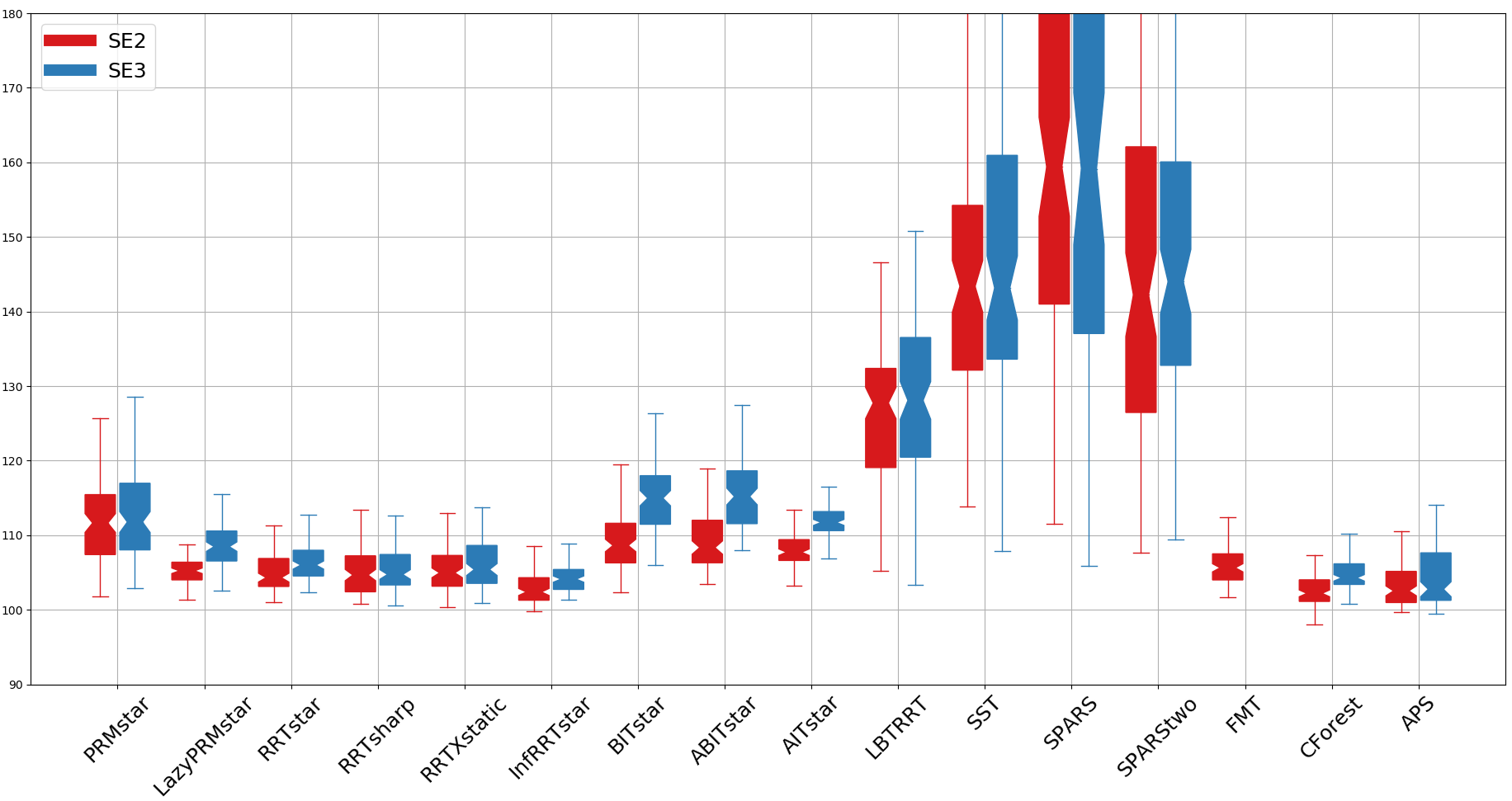}
  \caption{Resulting path lengths in SE2 and SE3 spaces, ground truth path lengths are normalized to 100 meters, all results are normalized accordingly}
  \label{fig:se2_se3_lengths}
\end{figure}

\begin{figure}
  \includegraphics[width=\linewidth]{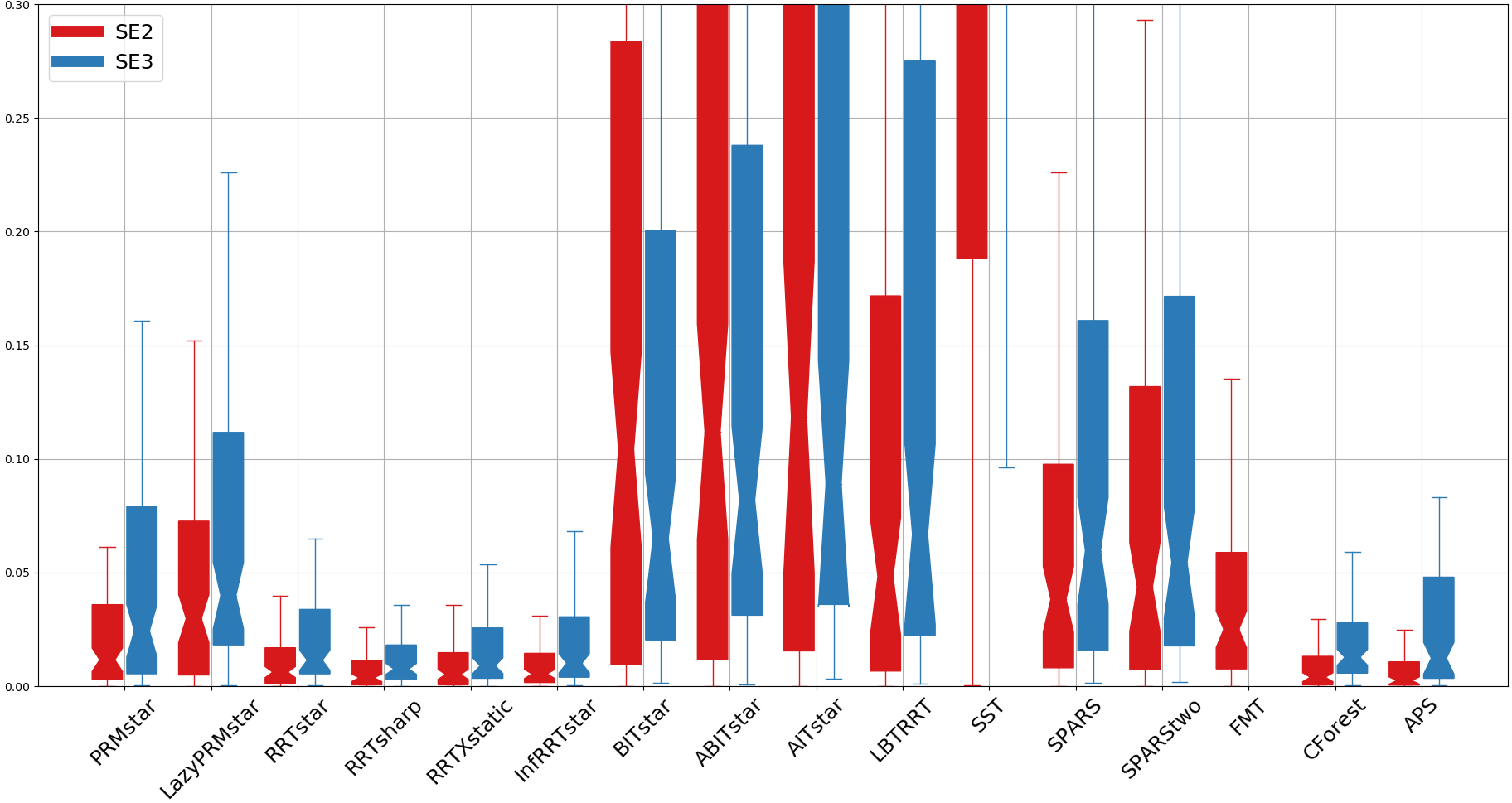}
  \caption{Box plot results of path smoothness in SE2 and SE3 space}
  \label{fig:se2_se3_smoothness}
\end{figure}

DUBINS and REEDS-SHEEP spaces are suitable while planing for car-like robots.
DUBINS state space particularly suits for forwards-only motions while REEDS-SHEEEP also allows backwards motions in generated plans. SE3 state space was considered for evaluating planners in non-flat environments, e.g. in figure \ref{fig:se3_result}.
In SE3, the resulting paths z axis can elevate/decrease within defined boundary. 
As can be seen from Figures \ref{fig:reeds_dubins_status} and \ref{fig:se2_se3_status}, planning in motion constrained space is much harder. Compared to SE2, the lengths and smoothness of resulting paths were not noticeably different, however one can see from figure \ref{fig:se2_se3_status} that the status of resulting paths differ due to increased state dimensions.

In general, planners perform better in relaxed SE2 and SE3 compared to DUBINS and REEDS-SHEEP in terms of providing a valid plan with given amount of time, this can be best seen from performance of FMT in figure \ref{fig:se2_se3_status} compared to figure \ref{fig:reeds_dubins_status}.
CFOREST and APS's parallelization approach seems to have a significant effect for quicker convergence towards optimal plans. 
In case of increased complexity in state spaces, as in DUBINS or SE3, FMT* planner was not able to provide a valid plans within required timeouts. In simpler cases though(e.g. SE2) FMT* was able to provide valid plans with average lengths and smoothness, see figures \ref{fig:se2_se3_lengths}, \ref{fig:se2_se3_smoothness} and \ref{fig:se2_se3_status}. 
SBO planners that are variants of RRT, performed dominantly better in terms of shorter and smoother plan generation in all state spaces.

\section{Conclusion} \label{conclusion}

In this work we evaluated the performances major state-of-the-art SBO planners in various state spaces with a randomized fashion. 
The evaluation setup was particularly aimed for planning of outdoor ground robots.
For convenience, OctoMaps were generated from simulation, the study faced two limitations, inability to do evaluation in larger scale environments(e.g. 1000 x 1000 meters) due to 
increased memory requirements and lack of suitable real data to be used as OctoMap. Despite these, the planners performances are little effected by the environment data, provided experimental results show strength/drawbacks of SBO planners in different use-cases. 
A general conclusion of this work can be drawn as, parallelized approaches such as CFOREST, AnytimePartShortening etc.
leads to consistency in generated plans and acceptable computation times(\textbf{1 second} for 85+ meter planning inquiries) for low speed ground robots. 

\bibliographystyle{plain}
\bibliography{main}
\end{document}